\documentclass[12pt,letterpaper]{article} 

\usepackage[margin=0.8in,includefoot=false,
            footskip=0.4in,nohead=true]{geometry}

\RequirePackage{url}
\usepackage[dvipsnames]{xcolor}

\usepackage[normalem]{ulem}
\usepackage{tikz}
\usepackage{longtable}
\usepackage[T1]{fontenc}
\usepackage{hyperref}
\usepackage{float}
\usepackage{setspace}
\usepackage{algorithm2e}
\usepackage{chapterbib}
\usepackage[numbers,sort&compress]{natbib}
\usepackage{verbatim}
\usepackage{authblk}
\usepackage[frak=mma,scr=rsfs,cal=mma]{mathalfa}
\DeclareMathAlphabet{\mathcal}{OMS}{cmsy}{b}{n}
\usepackage{tabularx}
\usepackage[margin=6pt,font={sf,footnotesize},labelfont=bf]{caption}
\usepackage{times}

\usepackage{amsmath}
\usepackage{amsfonts}
\usepackage{amssymb}
\usepackage{subfig}
\usepackage{dsfont}
\definecolor{linkcolour}{rgb}{0,0.2,0.6}
\hypersetup{colorlinks,breaklinks,linkcolor=linkcolour,citecolor=linkcolour,filecolor=linkcolour,urlcolor=linkcolour}
\usepackage{bm}
\usepackage{multirow} % merge rows in a table
\usepackage{booktabs}
\usepackage{arydshln}
\PassOptionsToPackage{hyphens}{url}\usepackage{hyperref}
\usepackage{enumitem}

\DeclareCaptionLabelSeparator{bar}{\;|\;}

\usepackage{sectsty}
\sectionfont{\large \sffamily\bfseries}
\subsectionfont{\normalsize \sffamily\bfseries}
\subsubsectionfont{\it \sffamily}

\makeatletter

\usepackage{titlesec}
\titlespacing*{\section}{0pt}{1.2\baselineskip}{0.8\baselineskip}
\titlespacing*{\subsection}{0pt}{0.8\baselineskip}{0.5\baselineskip}
\titlespacing*{\subsubsection}{0pt}{0.5\baselineskip}{0.3\baselineskip}

\newcommand{\model}{InstructPro\xspace}
\newcommand{\dataset}{InstructProBench\xspace}

\begin{document}

\title{\Large\sffamily\bfseries \model: Natural Language Guided Ligand-Binding Protein Design}

\normalsize
\author[1,*]{\sffamily Zhenqiao Song}
\author[2]{\sffamily Ramith Hettiarachchi}
\author[3]{\sffamily Chuan Li}
\author[3]{\\\sffamily Jianwen Xie}
\author[1,*]{\sffamily Lei Li}

\affil[1]{\small \sffamily Language Technologies Institute, School of Computer Science, \authorcr \small \sffamily Carnegie Mellon University, Pittsburgh, PA 15213, USA}
\affil[2]{\small \sffamily Ray and Stephanie Lane Computational Biology Department, School of Computer Science, \authorcr \small \sffamily Carnegie Mellon University, Pittsburgh, PA 15213, USA}
\affil[3]{\small \sffamily Lambda Inc, San Francisco, CA 94105, USA}
\affil[*]{\small \sffamily Correspondence:
\href{mailto:leili@cs.cmu.edu}{leili@cs.cmu.edu}
}
\date{\vspace{-5ex}}
\maketitle

\begin{abstract}
\noindent The \textit{de novo} design of ligand-binding proteins with tailored functions is essential for advancing biotechnology and molecular medicine, yet existing AI approaches are limited by scarce protein-ligand complex data. 
To circumvent this data bottleneck, we leverage the abundant natural language descriptions characterizing protein-ligand interactions. 
Here, we introduce \model, a family of generative models that design proteins following the guidance of natural language instructions and ligand formulas. 
\model produces protein sequences consistent with specified function descriptions and ligand targets. 
To enable training and evaluation, we develop \dataset, a large-scale dataset of 9.6 million (function description, ligand, protein) triples. 
We train two model variants --- \model-1B and \model-3B --- that substantially outperform strong baselines. 
\model-1B achieves an AlphaFold3 ipTM of 0.918 and a binding affinity of -8.764 on seen ligands, while maintaining robust performance in a zero-shot setting with scores of 0.869 and -6.713, respectively. These results are accompanied by novelty scores of 70.1\% and 68.8\%, underscoring the model's ability to generalize beyond the training set. Furthermore, the model yields a superior binding free energy of -20.9 kcal/mol and an average of 5.82 intermolecular hydrogen bonds, validating its proficiency in designing high-affinity ligand-binding proteins. Notably, scaling to \model-3B further improves the zero-shot ipTM to 0.882, binding affinity to -6.797, and binding free energy to -25.8 kcal/mol, demonstrating clear performance gains associated with increased model capacity.
These findings highlight the power of natural language-guided generative models to mitigate the data bottlenecks in traditional structure-based methods, significantly broadening the scope of \textit{de novo} protein design.

\end{abstract}

\section*{Introduction}
\label{introduction}
% 1. importance of ligand-binding protein design;
% 2. importance of utilizing text descriptions
% 3. our method (with one figure)
% experimental results & contributions

Proteins are fundamental to all forms of life and execute a wide range of biological functions. Many of these functions are mediated through specific protein-ligand interactions, such as catalyzing chemical reactions~\citep{kiss2013computational,rajakumara2022structure,dauparas2025atomic}, sensing small molecules that regulate signal transduction pathways~\citep{johnson2004trafficking,lee2016protein,wang2020small}, or recognizing cell-surface receptors to enable targeted drug delivery to diseased cells~\citep{vyas2001ligand,srinivasarao2017ligand,kunjiappan2021surface}. The ability to custom design ligand-binding proteins thus holds the promise of enabling a wide range of real-world applications, including therapeutics, diagnostics, industrial enzymes, biosensors, and molecular tools for synthetic and chemical biology~\citep{tinberg2013computational,yang2017computational,feng2017computational,bunzel2021designing,rottinghaus2022engineering,pennington2023harnessing}.

Recent advances in deep learning have led to remarkable progress in protein-ligand interaction prediction and \textit{de novo} design~\citep{ahmed2021deelig,chatterjee2023improving,rezaei2020deep,zhao2020exploring,rube2022prediction,dauparas2025atomic,ahern2025atom,butcher2025novo}. By enabling the direct design of proteins tailored to specific ligands, these methods have significantly reduced the dependence on resource-intensive, high-throughput experimental screening. Despite this progress, a fundamental challenge persists: the precise control of distinct function attributes beyond the singular goal of binding affinity. While recent generative models have introduced controllable design via keyword tags~\citep{nijkamp2023progen2}, prefix tokens~\citep{luo2023flexible}, or natural language prompts~\citep{guo2024protdat,praljak2024natural,xia2025naturelm}, a critical methodological disconnect remains. Existing language-guided frameworks generally lack the capacity for explicit conditioning on small-molecules, which is a prerequisite for the targeted engineering of ligand-binding proteins. Consequently, the integration of natural language guidance with ligand-aware protein design remains an under-explored area.

The development of such multimodal frameworks is further hindered by the limited protein-ligand complex structures~\citep{chakraborti2021all,siebenmorgen2024misato,cao2025surfdock}. High-resolution protein-ligand complex structures number only approximately 20,000, representing only a fraction of known small-molecule ligands. Conversely, ligand identities are annotated for millions of proteins in large-scale repositories like UniProtKB~\citep{uniprot2025uniprot} and a wealth of human-curated textual knowledge exists, describing the enzymatic mechanisms, binding specificities, and biophysical properties of proteins. These non-structural resources offer an underutilized opportunity to circumvent the limitations of structural data and enable large-scale ligand-binding protein design following human guidance.

% Furthermore, progress in traditional structure-based methods is constrained by the relatively small number of protein-ligand complex structures~(approximately 20,000), leaving substantial room for further exploration in this area~\citep{chakraborti2021all,siebenmorgen2024misato,cao2025surfdock}. Encouragingly, while structure data are scarce, ligand identities are annotated for many proteins in large-scale databases such as UniProtKB~\citep{uniprot2025uniprot}. Moreover, a wealth of human-curated textual knowledge exists, describing protein functions and ligand-binding properties. Together, these non-structural resources offer an underutilized opportunity to support large-scale ligand-binding protein design.

In this study, we introduce \model, a natural language-instructable~(e.g., Please design a protein which catalyzes phosphoribosyl pyrophosphate to diphosphate), multimodal framework for the \textit{de novo} design of ligand-binding proteins. \model comprises four key components: a text encoder, a ligand encoder, a shared memory module and a protein decoder. The text encoder encodes the combined human instruction and protein function description into comprehensive semantic representations, based on which a shared memory module extracts critical protein function information to facilitate more targeted and efficient subsequent design process. The ligand encoder processes ligand representations in SMILES format. Conditioned on both the extracted textual semantics and ligand representations, the protein decoder generates a protein sequence that aligns with the specified function description and is capable of binding the target ligand. This architecture enables \model to design ligand-binding proteins directly from textual instructions, without requiring structure input.

To summarize, our contributions are listed as follows:
\begin{itemize} [leftmargin=1.0em,topsep=0pt,parsep=0pt]
    \item We present \model, the first natural language instruction-following model and training approach for designing ligand-binding proteins directly from function descriptions and ligand formulas. 
    \item We curate \dataset, a large-scale dataset containing 9,592,829 (function description, ligand, protein) triples, enabling scalable training and evaluation for instruction-guided protein design.
    \item We develop two model variants, \model-1B (1 billion parameters) and \model-3B (3 billion parameters) that consistently outperform strong baselines. \model-1B achieves the highest AlphaFold3~\citep{abramson2024accurate} ipTM of 0.918 and binding affinity of -8.764 on seen ligands, and ipTM of 0.869, binding affinity of -6.713 and binding free energy of -20.9 kcal/mol in zero-shot setting, while \model-3B further improves these zero-shot scores to 0.882, -6.797 and -25.8 kcal/mol. These results demonstrate the effectiveness of natural language guidance in generating ligand-binding proteins. 
\end{itemize}

\section*{Our Approach}
\label{approach}
% \begin{figure*}
%   \centering
%   \includegraphics[width=15cm]{Figures/InstructPro_model.pdf}
%   \caption{\textbf{The overall architecture of \model.} The text encoder processes human instruction and protein function description in natural language. The shared memory module extracts essential contextual semantics from function description. The ligand encoder encodes molecular representations from the ligand SMILES formula, capturing the chemical context of the target ligand. Conditioned on both the critical contextual semantics and ligand representations, the protein decoder generates a protein sequence that both aligns with the function specification and is able to bind to the target ligand.}
%   \label{Fig: model}
% \end{figure*}

\begin{figure*}
  \centering
  \includegraphics[width=16.5cm]{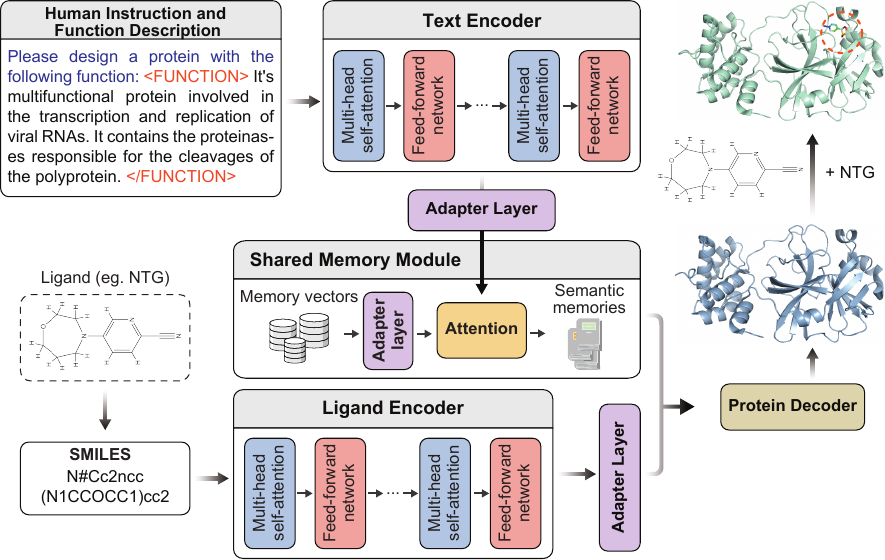}
  \caption{\textbf{The overall architecture of \model.} The text encoder processes human instruction and protein function description in natural language. The shared memory module extracts essential contextual semantics from function description. The ligand encoder encodes molecular representations from the ligand SMILES formula, capturing the chemical context of the target ligand. Conditioned on both the critical contextual semantics and ligand representations, the protein decoder generates a protein sequence that both aligns with the function specification and is able to bind to the target ligand.}
  \label{Fig: model}
\end{figure*}

\model is a natural language-instructable multimodal framework for the \textit{de novo} design of ligand-binding proteins.
The overall architecture is illustrated in Fig.~\ref{Fig: model}. \model comprises four core components: a text encoder, a ligand encoder, a shared memory module, and a protein decoder. The text encoder processes natural language input, including human instructions and protein function descriptions, to produce a sequence of text features. 
The shared memory module uses a fixed set of memory eliciting vectors to query text features and produce semantic memory representations. 
The ligand encoder captures the chemical context of the target ligand from its SMILES formula. Conditioned on both the semantic memory and ligand representations, the protein decoder generates a protein sequence that satisfies the specified functional and ligand-binding requirements. 

To train and evaluate \model, we construct \dataset, a large-scale dataset of (function description, ligand, protein) triples. We begin by extracting all natural language function annotations from UniProtKB~\citep{uniprot2025uniprot} and retain only those entries with recorded binding ligands. This results in 9,592,829 triples containing both function descriptions and associated ligands, corresponding to 7,535,394 unique UniProt entries (as some proteins are associated with multiple ligands).
To group proteins with similar sequences, we perform clustering using MMseqs2~\citep{steinegger2017mmseqs2} at a 30\% sequence identity rate, yielding 22,594 clusters. To ensure diversity in the validation and test sets, we stratify the clusters by size (number of sequences per cluster) and allocate: 20 clusters from the 1-500 sequence count range, 10 from 501-1000, 5 from 1001-2500, and 1 from clusters containing over 2500 sequences to each of the validation and test sets. The remaining clusters are used for training. This stratification ensures that the evaluation sets contain a mix of popular (large) and rare (small) protein sequence clusters. We further categorize the resulting examples based on whether their ligands are included in the training set. For validation, a set comprising 3,263 examples is constructed, of which the binding ligands do not present in the training set. For testing, we split 1,500 cases with seen ligands and 127 cases with ligands that are not observed in training set. The detailed dataset statistics are provided in Supplementary Table S1.

\begin{table*}[!t]
\small
\caption{Hyper-parameter settings of the \model family. For the text encoder, $N_t$, $\text{Head}_t$, and $D_t$ denote the number of layers, attention heads, and embedding dimensionality, respectively. For the ligand encoder, $N_g$, $\text{Head}_g$, and $D_g$ represent the same parameters. For the protein decoder, $N_p$, $\text{Head}_p$, and $D_p$ follow the same notation. $N_m$ denotes the memory eliciting vector size.}
\centering
\begin{tabular}{lccccccccccc}
\midrule
Models & Parameters & $N_t$ & $\text{Head}_t$ & $D_t$ & $N_m$ &  $N_g$ & $\text{Head}_g$ & $D_g$ & $N_p$ & $\text{Head}_p$ & $D_p$  \\
\midrule
\model-1B & 964M & 12 & 12 & 768 & 64 & 6 & 12 & 768 & 27 & 16 & 1,536 \\
\model-3B & 2.98B & 12 & 12 & 768 & 64 & 6 & 12 & 768 & 32 & 32 & 2,560 \\
\bottomrule
\end{tabular}
\label{Tab: model_specifications}
\end{table*}

Based on \dataset, we implement and train two model variants: \model-1B and \model-3B.
Both models use $N_t=12$ layers for text encoder with 12 attention heads and an embedding dimensionality $D_t = 768$. 
Its parameters are initialized with the pretrained PubMedBERT model~\citep{pubmedbert}.
Both models use $N_g=6$ layers for ligand encoder with 12 attention heads and an embedding dimensionality $D_g=768$. 
Its parameters are initialized with the pretrained SMILES-based RoBERTa model~\citep{ahmad2022chemberta}. 
We use $N_m=64$ memory eliciting vectors in the shared memory module, and a single layer feed-forward network in the adapter. 
\model-1B uses $N_p=27$ autoregressive Transformer~\citep{vaswani2017attention} decoding layers with 16 attention heads, and embedding dimensionality $D_p=1,536$. 
\model-3B uses $N_p=32$ with 32 attention heads and $D_p=2,560$.
The weights of both variants are initialized from ProGen2-base and ProGen2-BFD90~\citep{nijkamp2023progen2}, respectively. 
The total number of parameters is approximately 964 million for \model-1B and 2.98 billion for \model-3B. Table~\ref{Tab: model_specifications} summarizes the hyper-parameter settings of the \model family. 
Both \model-1B and \model-3B are trained for 1,000,000 steps using eight NVIDIA H100 GPUs. We adopt the Adam optimizer~\citep{kingma2014adam} with a learning rate warm-up over the first 10,000 steps, followed by linear decay. The learning rate is set to $1$e-$3$. The batch size is set to 16,384 tokens for \model-1B and 6,144 tokens for \model-3B, respectively. During inference, we employ greedy decoding to generate the output sequence. 
% The maximum sequence length is set to 1,024 residues.

\section*{Results}
\label{sec:results}

\subsection*{Baseline Models}
\label{subsection: baseline models}
We benchmark \model against two categories of baselines: natural language-guided models and structure-based models.
Natural language-guided baselines include:
(1) \textbf{Pinal}~\citep{dai2024toward}, an encoder-decoder framework that generates protein sequences conditioned on natural language descriptions of the targeted function.
(2) \textbf{ESM3}~\citep{hayes2025simulating}, a multimodal generative model capable of conditioning on function annotations expressed in natural language.
Structure-based baseline is:
(3) \textbf{AlphaFold3}+\textbf{LigandMPNN}~\citep{dauparas2025atomic}, which first predicts the complex structure of the ground-truth protein-ligand pair and then applies LigandMPNN to generate a sequence given the predicted structure.
Finally, since our protein decoder is initialized with ProGen2, we also compare against (5) \textbf{ProGen2-base}. To ensure a fair comparison, we generate protein candidates for all baseline models using their official codes. More baseline implementation details are provided in Supplementary Section 4.1.

\subsection*{Evaluation Metrics}
\label{subsection: evaluation metrics}
To evaluate whether the designed proteins exhibit the targeted function while maintaining structure stability, we assess them along three dimensions: functional performance, structure reliability and design novelty. 
Functional performance is conducted using: (1) \textbf{AlphaFold3 ipTM} which measures interface interaction geometry, with higher score indicating a more reliable binding. (2) \textbf{Binding Affinity} predicted by Gnina~\citep{mcnutt2021gnina} with initial poses generated from DiffDock~\citep{corso2024discovery}.
(3) \textbf{Binding Free Energy} calculated by MM/GBSA using AmberTools~\citep{case2023ambertools} with the molecular dynamics~(MD) simulation trajectory generated by OpenMM~\citep{eastman2023openmm}.
(4) \textbf{Number of Hydrogen Bonds} between the protein and ligand, which serves as a qualitative measure of their binding affinity.
Structure reliability is assessed by using: (5) \textbf{AlphaFold3 PAE} which evaluates the reliability of predicted structures, with lower values indicating greater folding reliability.
(6) \textbf{AlphaFold3 pLDDT} where scores above 80 suggest stable structures.
Finally, we compute (7) \textbf{Novelty} score of the designed proteins, calculated as ($1 - \text{amino acid recovery rate}$ after pairwise alignment with the ground truth sequence).

\begin{table*}[!t]
\small
\caption{Performance on seen ligands.}
\centering
\begin{tabular}{lccccc}
\midrule
Models & ipTM~($\uparrow$)  & Binding Affinity~($\downarrow$) &  PAE~($\downarrow$) & pLDDT ($\uparrow$)& Novelty~($\uparrow$)  \\
\cmidrule(r){1-6}
ProGen2~(764M) &	0.589&-7.377	&	10.496&	71.655&67.938\%\\
ESM3~(1.4B) &	0.624&	-7.846&10.181&	75.368&\textbf{78.416\%}\\
Pinal~(1.2B) &	0.751&	-7.526&	6.588&	81.419&70.118\%\\
AlphaFold3+LigandMPNN&	0.856&	-8.162&	2.371&	88.200& 61.560\%\\
\model-1B&\textbf{0.918}&	\textbf{-8.764}&\textbf{1.893}&	\textbf{89.601}	&70.146\%\\
\hdashline
\model-3B&0.877&-8.666&	2.261&	87.586&\textbf{71.880\%}\\
\bottomrule
\end{tabular}
\label{Tab: seen_ligands}
\end{table*}

\begin{table*}[!t]
\small
\caption{Performance on unseen ligands~(zero-shot setting).}
\centering
\begin{tabular}{lccccc}
\midrule
Models & ipTM~($\uparrow$)  & Binding Affinity~($\downarrow$)&  PAE~($\downarrow$) & pLDDT ($\uparrow$) & Novelty~($\uparrow$)  \\
\cmidrule(r){1-6}
ProGen2~(764M) & 0.686&-5.697&	9.079&	69.176	&50.950\%\\
ESM3~(1.4B) &0.774&-5.430&5.757&	86.857&\textbf{77.529\%}\\
Pinal~(1.2B) &	0.814&-6.071&	3.247&	87.157	&63.347\%\\
AlphaFold3+LigandMPNN&	0.868&	-6.325&	2.923&	89.862&67.291\%\\
\model-1B &\textbf{0.869}&\textbf{-6.713}&	\textbf{2.638}&	\textbf{90.518}&68.802\%\\
\hdashline
\model-3B&\textbf{0.882}&\textbf{-6.797}&\textbf{2.439}&	90.207&67.542\%\\
\bottomrule
\end{tabular}
\label{Tab: unseen_ligands}
\end{table*}

\begin{table*}[!t]
\small
\caption{Binding free energy and number of intermolecular hydrogen bonds between the protein and ligand on unseen ligand test set~(zero-shot setting).}
\centering
\begin{tabular}{lcc}
\midrule
Models & Binding Free Energy~(kcal/mol, $\downarrow$)  & Number of Hydrogen Bonds ($\uparrow$) \\
\cmidrule(r){1-3}
ProGen2~(764M) & -19.7185 & 5.32\\
ESM3~(1.4B) & -8.2438 & 2.66\\
Pinal~(1.2B) &	-16.1793 & 4.51\\
AlphaFold3+LigandMPNN& -19.7794	& 5.59\\
\model-1B & \textbf{-20.9063} & \textbf{5.82}\\
\hdashline
\model-3B& \textbf{-25.8011} & \textbf{6.45}\\
\bottomrule
\end{tabular}
\label{Tab: unseen_energy_hydrogen}
\end{table*}

\subsection*{Observations}
\label{subsection: main results}
% 1. compare instructpro with other baseline methods; 2. demonstrate zero-shot performance; 3. scaling up model performs better

The performance of our model on both seen and unseen ligands is summarized in Table~\ref{Tab: seen_ligands} and Table~\ref{Tab: unseen_ligands}, respectively. \model-1B demonstrates superior performance, yielding the highest scores across all function evaluation and structure reliability metrics. For function evaluation, \model-1B consistently achieves higher ipTM and binding affinity scores in both settings, validating its capability to design ligand-binding proteins with high affinity. Furthermore, it also obtains lower binding free energy and more number of intermolecular hydrogen bonds between the protein and ligand compared to baselines on the unseen test set. The better energetic scores and structural interactions validate the efficacy of \model in generating functional proteins with scaffolds with high ligand complementarity.
For structure reliability, the designs from \model-1B consistently surpass established thresholds, with PAE $< 10$ and pLDDT $> 80$~\citep{bennett2023improving, guo2022alphafold2}, confirming \model's ability to design proteins with stable and well-folded structures. While ESM3 exhibits slightly higher novelty scores, its significantly worse ipTM and binding affinity suggest a trade-off between its sequence divergence and function expression. In contrast, \model-1B maintains highly competitive novelty while effectively preserving targeted function, thereby validating its effectiveness in \textit{de novo} protein design.

\begin{figure*}
  \centering
  \includegraphics[width=17cm]{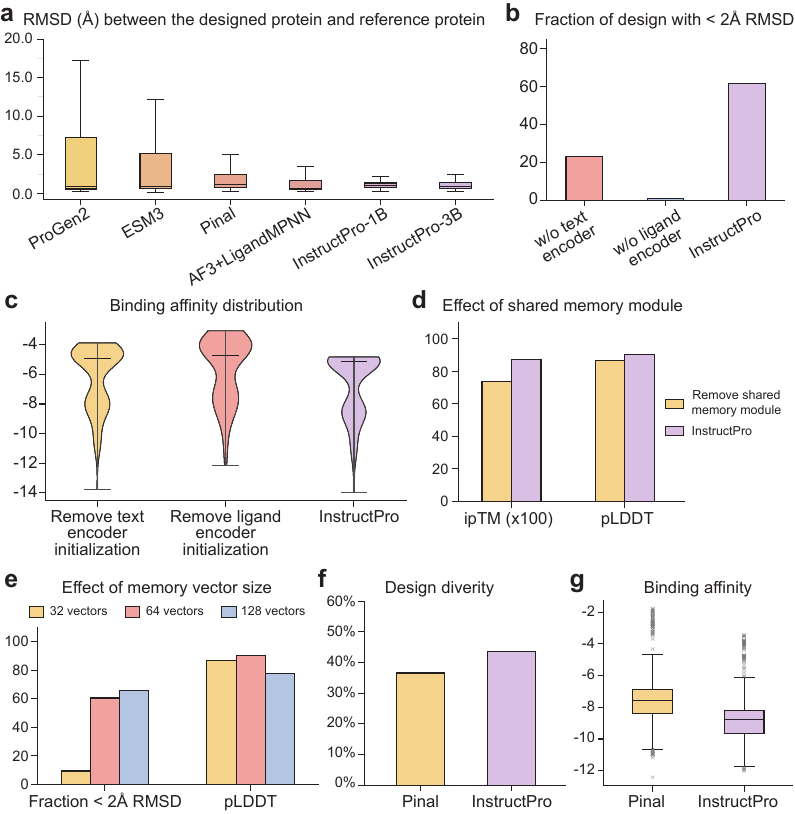}
  \caption{\textbf{Evaluation of \model.} \textbf{a,} The distribution of RMSD between the folded structure of designed proteins and ground truth proteins.
  \textbf{b,} The impact of removing~(w/o) text or ligand encoders. 
  \textbf{c,} The effect of applying pretrained text encoder or ligand encoder initialization.
  \textbf{d,} The influence of removing the shared memory module. \textbf{e,} The impact of memory eliciting vector size.
  \textbf{f,} The diversity score of Pinal and \model.
  \textbf{g,} The binding affinity distribution of Pinal and \model when applying sampling strategy.}
  \label{Fig: ablation_study}
\end{figure*}

To evaluate whether the designed proteins align with the function specifications in the natural language prompts, we analyze the RMSD distribution between the generated and ground truth proteins on seen ligands~(Fig. \ref{Fig: ablation_study}(a)). The results demonstrate that \model-1B outperforms all baseline methods, achieving the lowest RMSD and a ratio of 81.00\% for designs with an RMSD < 2Å. This high structure fidelity indicates that the generated proteins closely recapitulate the ground truth structures, confirming the \model's ability to translate textual function descriptions into a consistent protein sequence.

Scaling the model parameters from 1B to 3B yields additional performance gains, especially in unseen test set. On unseen ligands, the ipTM rises from 0.869 to 0.882, accompanied by an improvement in binding affinity from -6.713 to -6.797, binding free energy from -20.9 kcal/mol to -25.8 kcal/mol, average number of hydrogen bonds from 5.82 to 6.45 and PAE from 2.638 to 2.439. On seen ligands, the ratio of designs with RMSD < 2Å also increases from 81.00\% to 85.06\%. 
% Furthermore, the 3B variant also demonstrates stronger ability for \textit{de novo} design, with novelty scores increasing from 70.146\% to 71.880\% on seen ligands. 
These results confirm that \model adheres to established scaling laws, where improving model capacity directly translates into superior structure stability and function precision.

% \begin{figure*}
%   \centering
%   \includegraphics[width=17cm]{Figures/instructpro_results.pdf}
%   \caption{\textbf{Evaluation of \model.} \textbf{a,} The distribution of RMSD between the folded structure of designed proteins and ground truth proteins.
%   \textbf{b,} The impact of removing~(w/o) text or ligand encoders. 
%   \textbf{c,} The effect of applying pretrained text encoder or ligand encoder initialization.
%   \textbf{d,} The influence of removing the shared memory module. \textbf{e,} The impact of memory eliciting vector size.
%   \textbf{f,} The diversity score of Pinal and \model.
%   \textbf{g,} The binding affinity distribution of Pinal and \model when applying sampling strategy.}
%   \label{Fig: ablation_study}
% \end{figure*}

% \section*{Analysis}

% In this section, we provide a comprehensive analysis to demonstrate the effectiveness of \model, based on experiments conducted with \model-1B on the unseen ligand test set.

\subsection*{Ablation Study: Effect of Function Description and Ligand SMILES}
To quantify the contributions of function description and ligand SMILES to the design performance of \model, we conduct an ablation study by respectively removing the text encoder (\model-w/o text encoder) and ligand encoder (\model-w/o ligand encoder). Evaluation on the unseen test set (Fig. \ref{Fig: ablation_study}(b)) reveals that the absence of either module leads to a significant decrease in the function score, measured by the ratio of designs achieving an RMSD < 2Å. Notably, the removal of the ligand encoder causes a greater decline. These findings demonstrate that both textual function descriptions and chemical formula information are critical for the accurate design of proteins with targeted functions.

We further investigate the influence of the model weight initialization from pretrained encoders. Binding affinity distributions~(Fig. \ref{Fig: ablation_study}(c)), predicted by Gnina with initial positions generated from DiffDock, indicate that the absence of pretraining negatively impacts the binding affinity between the designed proteins and the target ligands. The effect is more pronounced after removing the ligand encoder, leading to a substantial binding affinity drop. These results verify the importance of leveraging pretrained representations to effectively capture the complex biophysical constraints essential for targeted ligand-binding functions.

\subsection*{Ablation Study: Impact of Shared Memory Module}
To evaluate the impact of the shared memory module on model performance, we develop an ablation variant in which this component is removed and the full set of raw textual semantic features is utilized. The resulting AlphaFold3 ipTM score and pLDDT value on the unseen ligand test set are shown in Fig.~\ref{Fig: ablation_study}(d). The results reveal that the removal of the shared memory module leads to a decline in both function exhibition and structure stability. These findings confirm the module's essential role in distilling key information from textual features, thereby facilitating more functionally consistent protein design with their natural language specifications.

To explore the influence of memory eliciting vector size, we further train two variants with 32 and 128 memory vectors, respectively. As shown in Fig.~\ref{Fig: ablation_study}(e), reducing the size to 32 results in significant performance degradation on the unseen ligands. Increasing it to 128 slightly improves the function score (\% < 2Å RMSD) from 60.62\% to 66.14\%, but substantially decreases structure stability, with pLDDT dropping from 90.518 to 77.855. To balance the design quality~(function and structure stability), we finally adopt a  memory eliciting vector size of 64.

\subsection*{\model Is Able to Design Diverse Proteins}
To evaluate the diversity of generated proteins, we apply Nucleus Sampling~\citep{holtzmancurious} with a probability threshold of $p=0.4$ on the seen ligand test set. For each function description and ligand SMILES, we generate five candidate sequences, among which we compute the diversity rate for each candidate pair as $1 -$ (amino acid recovery rate after pairwise alignment) and finally report the average diversity rate. As shown in Fig.~\ref{Fig: ablation_study}(f-g), \model achieves a diversity score of 42.79\%, surpassing Pinal's 33.56\%. At the same time, proteins designed by \model also obtain better binding affinity. These observations demonstrate that \model is capable of generating diverse proteins which maintain strong binding to the specified ligands.

\begin{figure*}
  \centering
  \includegraphics[width=17cm]{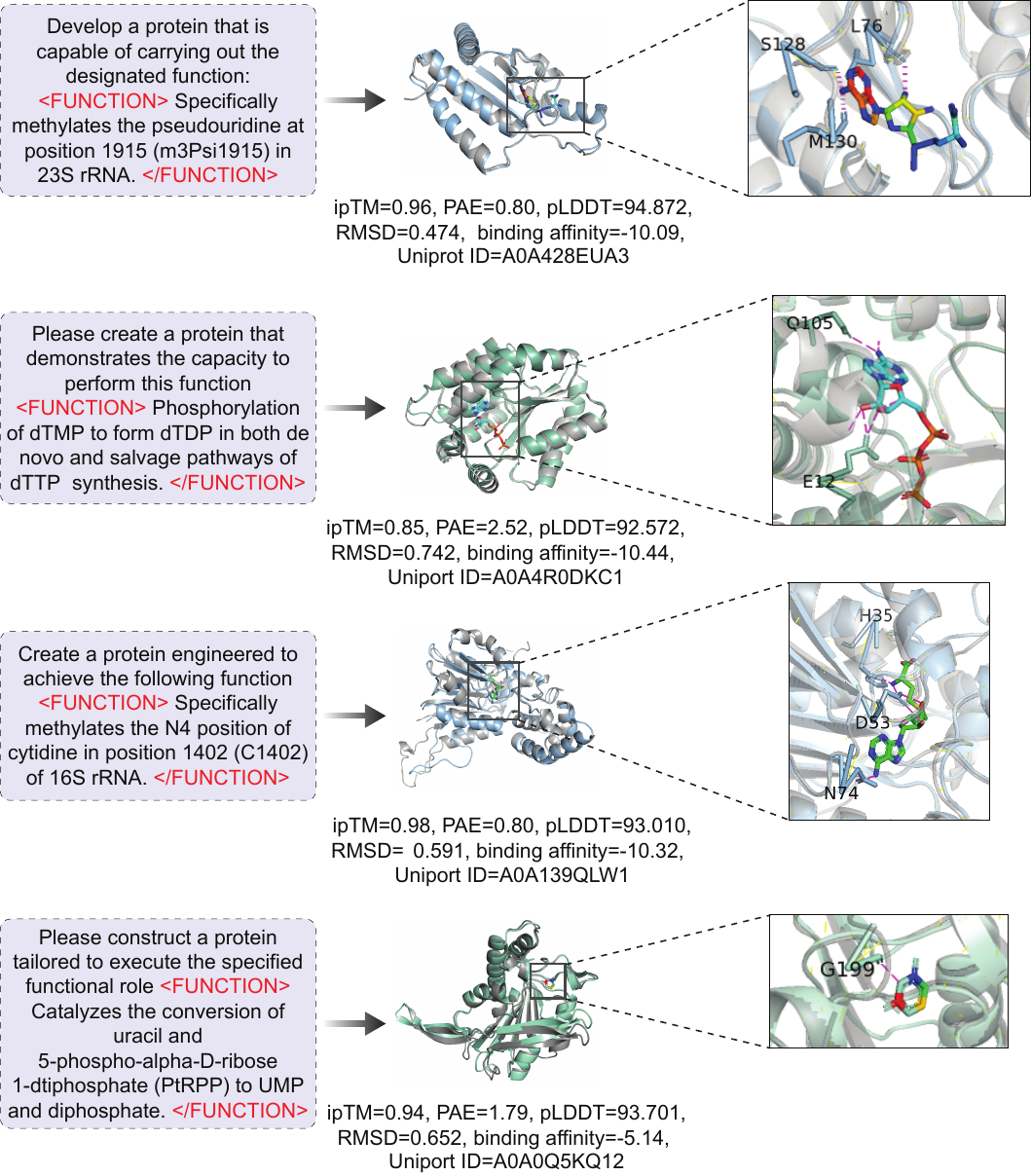}
  \caption{\textbf{Ligand-binding proteins designed by \model-1B.} In all examples, natural proteins are depicted in grey, while the proteins designed by \model-1B are shown in green or blue. The hydrogen bonds between the designed protein and the target ligand are shown in purple.}
  \label{Fig: case_study}
\end{figure*}

\subsection*{Case Study: Designing Ligand-Binding Proteins with \model}
Figure~\ref{Fig: case_study} illustrates four representative ligand-binding proteins generated by \model-1B, sampled from both seen and unseen ligand test sets. Across all cases, the designs consistently achieve good performance on all metrics, including ipTM scores > 0.8, PAE < 5, pLDDT > 90, RMSD < 2Å with high binding affinity. These findings demonstrate a good alignment  between the function of the generated proteins and the natural language prompts, validating \model's ability to design structurally reliable proteins that precisely recapitulate user-specified functional intent.

\section*{Discussion}
In this study, we introduce \model, a natural language-instructable, multimodal framework for designing ligand-binding proteins. The framework consists of four core components: a text encoder, a ligand encoder, a shared memory module, and a protein decoder. Given a textual description of the desired function and a ligand SMILES formula, \model generates a protein sequence that functionally align with the specified instruction and capable of binding to the target ligand with high affinity. We develop and evaluate two variants, \model-1B and \model-3B, containing 1 billion and 3 billion parameters, respectively. Both consistently outperform strong baselines across multiple evaluation metrics. Notably, \model-1B achieves the highest ipTM score, binding affinity, pLDDT and PAE on both seen and unseen ligand test sets while maintaining high design novelty. 
Scaling to 3B parameters leads to further performance gains, verifying \model is scalable. We envision \model will serve as a useful tool in efficient protein design, enabling the creation of a new generation of small-molecule ligand-binding proteins and bridging the gap between human functional intent and resulted protein.

% limitations and further improvement
While our work demonstrates superior performance comparing to current state-of-the-art methods and enables the \textit{de novo} generation of ligand-binding proteins from natural language, several avenues for future development remain. 
First, due to computational constraints, the current \model is trained exclusively on ligand-binding proteins, a subset of the broader protein database UniProt (10 million versus 200 million). To enhance the model's generalizability and generative robustness, future work can consider a two-stage training paradigm:  (1) pre-training on the full UniProt database to capture universal evolutionary principles, followed by (2) continuing training on our curated \dataset. This hierarchical approach is expected to significantly broaden the model's understanding of protein folds and improve its capacity for \textit{de novo} design across diverse families.
Second, while the design of small-molecule ligand-binding proteins is critical for wide applications like biosensing and biocatalysis, the current scope of \model is limited to this function domain. Expanding the framework to accommodate a wider array of human instructions, such as protein-protein interactions, cellular localization, and thermal stability, would significantly broaden its utility. Developing a truly general-purpose, instructable protein design tool will necessitate the integration of broader multimodal datasets and more complex instruction-following architectures to address a more diverse range of biotechnological challenges.
Third, while we have rigorously evaluated our designs using multiple \textit{in silico} function evaluation metrics, including AlphaFold3 ipTM, structure RMSD, binding affinity and binding free energy, our results currently lack experimental validation. Although these computational metrics are strongly correlated with physical viability, the incorporation of wet-lab validation will definitively increase the reliability of the designed proteins.

In summary, this study establishes a robust paradigm for the \textit{de novo} design of proteins guided by natural language instructions. Our results provide a foundation for next-generation generative models that promise to achieve \textit{de novo} design of small-molecule ligand-binding proteins by directly translating human-specified function requirements into desired proteins.

\section*{Methods}
\label{methods}
\subsection*{Problem Definition}

% A protein is composed of a linear chain of amino acids, covalently linked by peptide bonds and folded into a specific 3D structure. 
Let $\mathcal{A}$ denote the set of the 20 standard amino acids. We represent a protein sequence of length $N_r$ as $\boldsymbol{r} = \{r_1, r_2, \ldots, r_{N_r}\} \in \mathcal{A}^{N_r}$, where each residue $r_i \in \mathcal{A}$ for $i \in \{1, \ldots, N_r\}$. 
To guide the design, we concatenate a user-provided instruction and a corresponding protein function description into a natural language sequence $\boldsymbol{w} = \{w_1, w_2, \ldots, w_{N_w}\}$ of length $N_w$, where the function description is marked with special tokens \texttt{<FUNCTION>} and \texttt{</FUNCTION>}. $w_j \in \mathcal{V}$ for $j\in \{1, 2, ..., N_{w}\}$ and $\mathcal{V}$ denotes the natural language token vocabulary. Each ligand is represented by a SMILES string $\boldsymbol{s} = \{s_1, s_2, \ldots, s_{N_s}\}$ with length $N_s$, and each token $s_k \in \mathcal{S}$ for $k\in \{1, 2, ..., N_{s}\}$, with $\mathcal{S}$ denoting the SMILES token alphabet.  

The task is formulated as follows: given a user-provided instruction and protein function description $\boldsymbol{w}$, and a target ligand $\boldsymbol{s}$, generate a protein sequence $\boldsymbol{r}$ that binds to the ligand and satisfies the described function. Formally, we aim to learn a conditional generative model $P(\boldsymbol{r} \mid \boldsymbol{w}, \boldsymbol{s}; \theta)$, parameterized by $\theta$.

\subsection*{Text Encoder and Shared Memory Module}
The text encoder encodes the combined human instruction and protein function description into comprehensive semantic representations. It consists of a $N_{t}$-layer Transformer encoder. Each layer comprises a multi-head self-attention (MHA) sub-layer and a feed-forward network (FFN), both followed by residual connections and layer normalization. The representation of each token in the input text is computed as:
\begin{equation}
\small
\begin{split}
\boldsymbol{h}^0_{t,i}&=\boldsymbol{E}_{t}\cdot \text{onehot}(w_i)\\
\Tilde{\boldsymbol{h}}_{t,i}^{l+1}&=\mathrm{LayerNorm}\left(\mathrm{MHA}(\boldsymbol{h}_{t,i}^l, \boldsymbol{H}_t^l)+\boldsymbol{h}_{t,i}^l; \theta_t \right)\\
\boldsymbol{h}_{t,i}^{l+1}&=\mathrm{LayerNorm} \left(\mathrm{FFN}(\Tilde{\boldsymbol{h}}_{t,i}^{l+1})+\Tilde{\boldsymbol{h}}_{t,i}^{l+1};\theta_t \right) 
\end{split}
\end{equation}
where $\boldsymbol{h}_{t,i}^{l}$ is the $i^{th}$ token input representation at the $l^{\mathrm{th}}$ layer and $\boldsymbol{h}_{t,i}^{0}$ is the $i$-th token embedding in the text sequence. $\boldsymbol{H}_t^l=[\boldsymbol{h}_{t,1}^l, \boldsymbol{h}_{t,2}^l, ..., \boldsymbol{h}_{t,N_w}^l]^T$ denotes the full sequence input at layer $l$. The embedding matrix $\boldsymbol{E}_t$ maps each input token $w_i$ into a dense vector of dimensionality $D_t$. All parameters in the text encoder are denoted by $\theta_t$.

To mitigate the inefficiency introduced by long textual function descriptions which may span thousands of tokens, we
introduce a shared memory module to elicit key semantic information. 
Specifically, we introduce a set of $N_m$ trainable memory eliciting vectors $\boldsymbol{M}$, which serve as queries in an attention mechanism to extract essential textual features from the final layer output representations of the text encoder:
\begin{equation}
\small
\hat{\boldsymbol{H}}_t=\text{Softmax}(W_q\boldsymbol{M}\cdot W_k\boldsymbol{H}_t^{N_t})W_v\boldsymbol{H}_t^{N_t}
\end{equation}
% \begin{equation}
% \small
% \hat{\boldsymbol{H}}_t=\text{Single-Head-Attention}(\boldsymbol{M}, \boldsymbol{H}_t)
% \end{equation}
where $W_q$, $W_k$, and $W_v$ are learnable projection matrices with dimensionality $D_t \times D_t$. This mechanism enables the model to selectively retain and compress critical semantic content from the input instructions and function descriptions, facilitating more targeted and efficient protein sequence generation.

\subsection*{Ligand Encoder}
To model the binding-ligand information, we employ another Transformer encoder with $N_{g}$ layers to encode the ligand's SMILES strings. The resulting contextual representations capture structural and chemical properties of the ligand, providing essential cues to guide the generation of protein sequences capable of binding the target ligand.

Specifically, the encoding of each token in the SMILES string is calculated as:
\begin{equation}
\small
\begin{split}
\boldsymbol{h}^0_{g,i}&=\boldsymbol{E}_{g}\cdot \text{onehot}(s_i) \\
\Tilde{\boldsymbol{h}}_{g,i}^{l+1}&=\mathrm{LayerNorm}\left(\mathrm{MHA}(\boldsymbol{h}_{g,i}^l, \boldsymbol{H}_g^l)+\boldsymbol{h}_{g,i}^l; \theta_g \right)\\
\boldsymbol{h}_{g,i}^{l+1}&=\mathrm{LayerNorm} \left(\mathrm{FFN}(\Tilde{\boldsymbol{h}}_{g,i}^{l+1})+\Tilde{\boldsymbol{h}}_{g,i}^{l+1};\theta_g \right), 
\end{split}
\end{equation}
where $\boldsymbol{h}_{g,i}^{l}$ denotes the input representation of the $i^{th}$ SMILES token at the $l^{\mathrm{th}}$ layer and $\boldsymbol{h}_{g,i}^{0}$ is the $i$-th SMILES token embedding. $\boldsymbol{H}_g^l=[\boldsymbol{h}_{g,1}^l, \boldsymbol{h}_{g,2}^l, ..., \boldsymbol{h}_{g,N}^l]^T$ represents the full sequence input at layer $l$. The embedding matrix $\boldsymbol{E}_g$ maps one-hot encoded token $s_i$ to continuous vectors of dimensionality $D_g$. $\theta_g$ denotes the trainable parameters in the ligand encoder. 

The ligand representations from the final layer are subsequently integrated with the semantic memory vectors, serving as conditioning inputs to the protein decoder. This fusion ensures that the generated protein sequence reflects both the functional intent and the structure requirements necessary for ligand binding.

\subsection*{Protein Decoder}
The protein decoder is designed to generate a protein sequence that both aligns with the given protein function description and is able to bind to the specified ligand.
Specifically, we adopt a Transformer decoder-based architecture~\citep{vaswani2017attention} to autoregressively generate the protein sequence. At every decoding step, prefix tuning~\citep{li2021prefix} is applied to get the conditional probability distribution of each residue $r_i$. The overall model contains $N_p$ layers of unidirectional Transformer layer, with embedding size $D_p$:
% \begin{equation}
% \small
% \begin{split}
% \hat{\boldsymbol{H}}_g&= \mathrm{FFN}(\boldsymbol{H}_g^{N_g})\\
% \boldsymbol{h}_{p,i}&=F_\text{protein}(\boldsymbol{r}_{< i}| \hat{\boldsymbol{H}_t}, \hat{\boldsymbol{H}}_g;\theta_p)\\
% P(r_i) &= \text{Softmax}(\boldsymbol{h}_{p, i})
% \end{split}
% \end{equation}
\begin{equation}
\small
\hat{\boldsymbol{H}}_g= \mathrm{FFN}(\boldsymbol{H}_g^{N_g}),\;\boldsymbol{h}_{p,i}=F_\text{protein}(\boldsymbol{r}_{< i}| \hat{\boldsymbol{H}_t}, \hat{\boldsymbol{H}}_g;\theta_p),\;P(r_i) = \text{Softmax}(\boldsymbol{h}_{p, i})
\end{equation}
where $F_{\text{protein}}$ denotes the protein decoder network parameterized by $\theta_p$, and $\boldsymbol{r}_{< i}=\{r_1, r_2, ..., r_{i-1}\}$ is the partial sequence generated before position $i$. The vectors $\hat{\boldsymbol{H}}_t$ and $\hat{\boldsymbol{H}}_g$ represent the semantic memory vectors and the final chemical contexts from the ligand inputs, respectively. The adapter module $\mathrm{FFN}$ projects the ligand features from the ligand space into the residue representation space, ensuring modality alignment.

The model is trained to minimize the negative log-likelihood of the target protein sequence:
\begin{equation}
\small
\mathcal{L}=\sum_{i=1}^{N_r} -\log P(r_i)
\end{equation}
where $N_r$ denotes the length of the designed protein sequence. The prefix tuning strategy encourages the protein decoder to generate protein sequences that are both functionally consistent with the textual description and structurally complementary with the target ligand.

\section*{Data Availability}
\label{data_link}
\sloppy
All datasets utilized in this study are publicly accessible. Functional annotations, recorded binding ligands, and corresponding protein sequences are systematically mined from the UniProtKB database (\url{https://www.uniprot.org/uniprotkb}). Ligand SMILES strings are sourced from the ChEBI repository (\url{https://ftp.ebi.ac.uk/pub/databases/chebi/SDF/}). The final, curated dataset \dataset is available at \href{https://drive.google.com/file/d/1u5L8gJ1WEMzPg0b4NpzSWUsxYoXmZyK4/view?usp=sharing}{data}.

\section*{Code Availability}
\sloppy
The code and model checkpoints used in this study are publicly available. The most updated and supported codebase is located at \href{https://github.com/JocelynSong/InstructPro.git}{\model codebase}.

\clearpage
\bibliographystyle{naturemag}
\bibliography{refs}

\clearpage
\section*{Acknowledgment}
This work was supported, in part, by an NEC Faculty Research Award and the Neocortex Award from the Pittsburgh Supercomputing Center.

\section*{Author Contributions}
Conceptualization, Z.S., L.L.; 
Methodology, Z.S., L.L.; 
Software, Z.S., R.H., L.L.;
Investigation, Z.S., R.H., J.X., L.L.; 
Writing, Z.S., R.H., L.L.; 
Funding Acquisition, C.L., J.X., L.L.

\section*{Competing Interests}
The authors declare no competing interests.

\clearpage
\appendix 

\setcounter{page}{1}
\setcounter{section}{0}
\renewcommand{\thesection}{\arabic{section}} % Sections labeled as A, B, C, ...
\renewcommand{\thesubsection}{\thesection.\arabic{subsection}} % Subsections as A.1, A.2, ...
\renewcommand{\thesubsubsection}{\thesubsection.\arabic{subsubsection}} % Subsubsections as A.1.1, A.1.2, ...

% Number figures and tables within the appendix sections
\renewcommand{\thetable}{S\arabic{table}}
\setcounter{table}{0}

\renewcommand{\thefigure}{S\arabic{figure}}
\setcounter{figure}{0}

\begin{center}
{\Large\sffamily \bfseries \model: Natural Language Guided Ligand-Binding Protein Design \\
\vspace{20pt} (SUPPLEMENTAL INFORMATION)}
\end{center}

\section{Related Work}
\label{related_work}
\textbf{Ligand-Binding Protein Design.} 
Designing proteins that bind to specific ligands requires known information about the key interacting residues. Therefore, most work in this area has focused on redesigning proteins based on known structural motif information \citep{hellinga1991construction, tinberg2013computational, dou2017sampling,nguyen2024proteinrediff}. However, recent advances in \textit{de novo} protein design have enabled the generation of proteins tailored to specific ligands \citep{krishna2024generalized, lu2024novo, zambaldi2024novo,ahern2025atom}. For instance, generative models such as RFDiffusionAA \citep{krishna2024generalized} can produce novel protein backbones tailored to bind a specific ligand, often by building around a defined ligand pose. Once such candidate proteins are designed, they can be further screened by molecular dynamics simulations \citep{barros2019improving}, docking \citep{corso2024discovery} or by using co-folding methods \citep{krishna2024generalized, abramson2024accurate}.
Despite the advancements in such \textit{de novo} design methods, precisely controlling for other specific functional attributes of the protein, beyond just binding the ligand, remains a significant area for development \citep{chu2024sparks}.
% adding RFdifusion2 into the related work

\textbf{Natural Language Guided Protein Design.}
Recently, there has been a shift from controlled protein design using methods such as keyword tags \citep{nijkamp2023progen2,hayes2025simulating}, learned prefix tokens \citep{luo2023flexible} or task specific instructions \citep{lv2025prollama} to make up much more general language-guided protein design methods \citep{guo2024protdat, dai2024toward,praljak2024natural,riley2025generalized,xia2025naturelm,liu2025text}. For instance, BioM3 \citep{praljak2024natural} generates protein sequences through an auto-regressive diffusion model. \cite{dai2024toward} proposes Pinal, a model that generates three-dimensional~(3D) structure tokens first and then generates protein sequences. They find that such two-stage approach helps to limit the search space, rather than end-to-end text-to-sequence generation.  Furthermore, \cite{riley2025generalized} proposes a transformer-based model (MP4) for end-to-end text-to-sequence generation, and have performed experimental validation to test the ability of the generated protein sequences to stably express. Although these methods demonstrate the capability of natural language-guided protein design, they do not incorporate conditioning on small molecules, which is necessary, for example, in designing ligand-binding proteins. 

Accounting for the limitations in both of these use cases, we propose InstructPro, a language-guided ligand-binding protein design model that goes beyond understanding just protein sequences.

\section{The Use of Large Language Models }
In this work, large language models are only used for typo correction and grammar checking.

\section{Dataset Details}

We provide detailed data statistics of our curated \dataset in Table \ref{tab:data_statistics}.

\begin{table*}[ht]
\setlength{\tabcolsep}{1.0mm}
\caption{Detailed data statistics for \dataset.}
\centering
\begin{tabular}{lccccc}
    \toprule
Split       & Size    & Unique Uniprot ID & Unique Function Description & Unique Ligand \\
\midrule
Training       & 9,586,972  &      7,530,857  &     11,271   &       248       \\
Validation         &    3,263   &        1,943    &         41   &         6       \\
Test (Seen Ligand)  &    1,500   &        1,500    &        195   &        11       \\
Test (Unseen Ligand) &    127   &        127    &    43        &         5       \\
    \bottomrule
\end{tabular}
\label{tab:data_statistics}
\end{table*}

\section{Additional Experimental Information}

\subsection{Additional Baseline Implementation Details}
\label{sec:baseline_implementation_details}
We provide further implementation details of baseline methods as follows:
\begin{itemize}
\item \textbf{Pinal}: We follow the instructions on their official GitHub repository to achieve design.
\item \textbf{ESM3}: We begin by extracting functional annotations for each protein from InterPro~\citep{blum2025interpro}. Using these annotations as conditioning signals, we apply ESM3 to generate an initial protein sequence and structure. Next, conditioned on both the functional annotation and the designed sequence, we refine the protein structure. Finally, leveraging the improved structure, we design a corresponding protein sequence. For evaluation, we employ the publicly available ESM3-open model, which contains 1.4 billion parameters.
\item \textbf{AlphaFold3}+\textbf{LigandMPNN}:
We first get the complex structure of the target ligand and ground truth protein using AlphaFold3. Then taking the ligand structure and protein backbone structure from the folded complex as model input, LigandMPNN correspondingly designs a protein sequence.
\item \textbf{ProGen2}: Since ProGen2 is an autoregressive protein language model without any conditioning signals, we provide the first 32 amino acids as a prefix to guide the model design.
\end{itemize}

\clearpage
\bibliographystyle{naturemag}
\bibliography{refs}
\clearpage

\end{document}